\documentclass[lettersize,journal]{IEEEtran}
\usepackage[T1]{fontenc}
\usepackage{graphicx}
\usepackage[ruled,linesnumbered]{algorithm2e}
\usepackage{booktabs}
\usepackage{amssymb}
\usepackage{multicol}
\usepackage{multirow}
\usepackage{threeparttable}
\usepackage{cite}
\usepackage{url}

\usepackage{graphics} 
\usepackage{amsmath} 
\usepackage[export]{adjustbox}
\usepackage[table,xcdraw]{xcolor}
\usepackage{stfloats}

\begin{document}

\title{Extrinsic Manipulation on a Support Plane by Learning Regrasping}
\author{
Peng Xu$^{1}$, Zhiyuan Chen$^{1}$, Jiankun Wang$^*$,~\IEEEmembership{Senior Member,~IEEE}, and Max Q.-H. Meng$^*$,~\IEEEmembership{Fellow,~IEEE}
\thanks{$^{1}$Peng Xu and Zhiyuan Chen contributed equally to this work. $^*$Corresponding authors: Jiankun Wang; Max Q.-H. Meng.}
\thanks{Peng Xu is with the Department of Electronic Engineering, The Chinese University of Hong Kong, Hong Kong
        {\tt\small peterxu@link.cuhk.edu.hk}}%
\thanks{Zhiyuan Chen and Jiankun Wang are with Shenzhen Key Laboratory of Robotics Perception and Intelligence, and the Department of Electronic and Electrical Engineering, Southern University of Science and Technology, Shenzhen 518055, China
        {\tt\small 12132111@mail.sustech.edu.cn, \tt\small wangjk@sustech.edu.cn}}%
\thanks{Max Q.-H. Meng is with Shenzhen Key Laboratory of Robotics Perception and Intelligence and the Department of Electronic and Electrical Engineering at Southern University of Science and Technology in Shenzhen, China. He is a Professor Emeritus in the Department of Electronic Engineering at The Chinese University of Hong Kong in Hong Kong and was a Professor in the Department of Electrical and Computer Engineering at the University of Alberta in Canada.
        {\tt\small max.meng@ieee.org}}%
}

\markboth{Journal of \LaTeX\ Class Files,~Vol.~0, No.~0, February~2022}%
{Shell \MakeLowercase{\textit{et al.}}: A Sample Article Using IEEEtran.cls for IEEE Journals}

\IEEEpubid{0000--0000/00\$00.00~\copyright~2022 IEEE}

\maketitle

\begin{abstract}

Extrinsic manipulation, a technique that enables robots to leverage extrinsic resources for object manipulation, presents practical yet challenging scenarios. Particularly in the context of extrinsic manipulation on a supporting plane, regrasping becomes essential for achieving the desired final object poses. This process involves sequential operation steps and stable placements of objects, which provide grasp space for the robot. 
To address this challenge, we focus on predicting diverse placements of objects on the plane using deep neural networks. A framework that comprises orientation generation, placement refinement, and placement discrimination stages is proposed, leveraging point clouds to obtain precise and diverse stable placements. To facilitate training, a large-scale dataset is constructed, encompassing stable object placements and contact information between objects. Through extensive experiments, our approach is demonstrated to outperform the start-of-the-art, achieving an accuracy rate of 90.4\% and a diversity rate of 81.3\% in predicted placements. Furthermore, we validate the effectiveness of our approach through real-robot experiments, demonstrating its capability to compute sequential pick-and-place steps based on the predicted placements for regrasping objects to goal poses that are not readily attainable within a single step. Videos and dataset are available at https://sites.google.com/view/pmvlr2022/. 



\end{abstract}



\begin{figure}[htp]
      \centering
      \includegraphics[width=8.5cm]{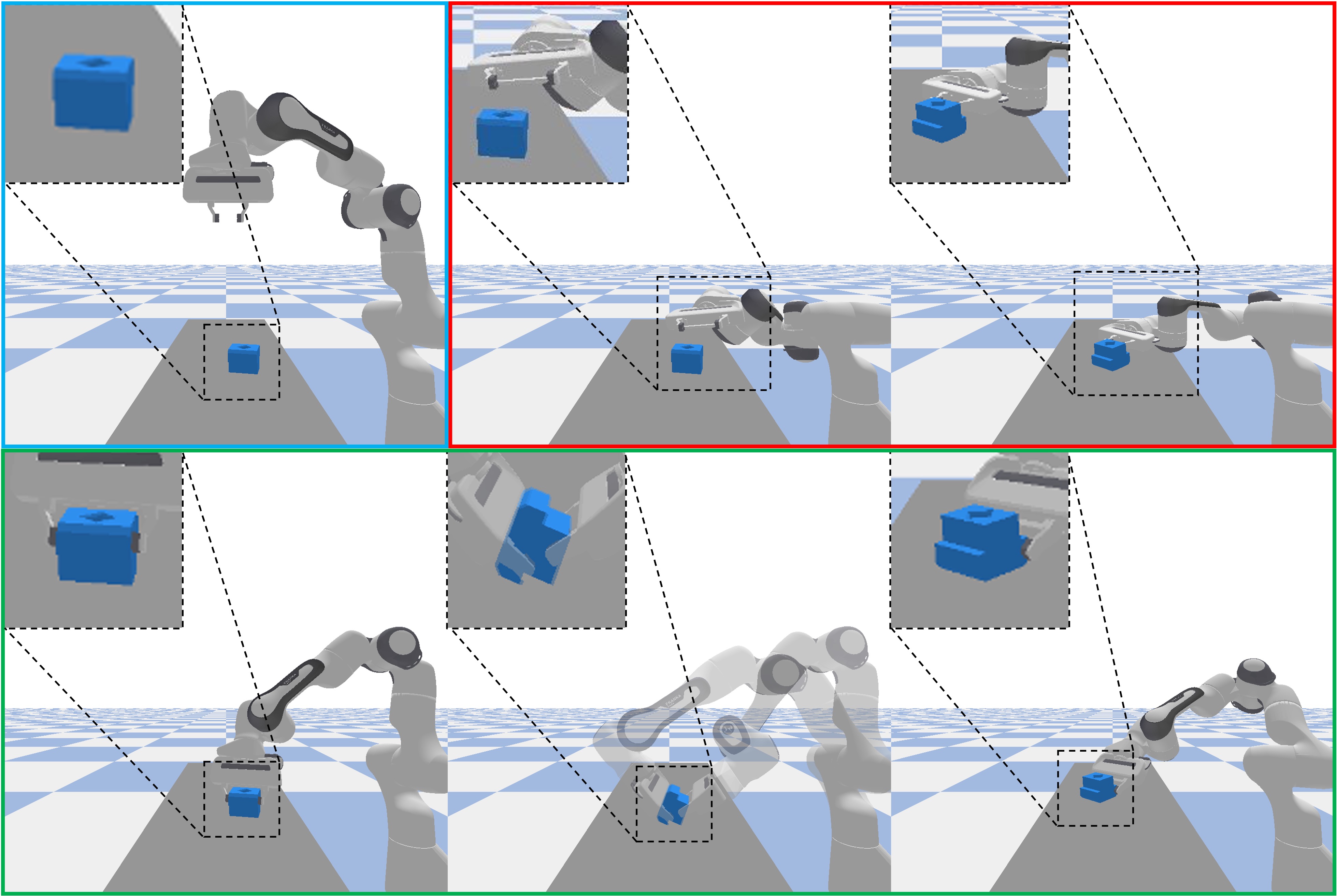}
      \caption{An example of flipping a T-nut. The robot can not flip the object in its initial state (blue border) with one pick-rotate-and-place step (red border) due to kinematic constraints and environmental occlusion. However, based on the stable placements predicted by our framework, the robot can flip the object with sequential steps (green border).}
      \label{f0}
   \end{figure}

\section{Introduction}

\IEEEPARstart{E}{xtrinsic}
 manipulation refers to leveraging resources extrinsic to the gripper for object manipulation \cite{6907062}. An extrinsic supporting plane, for example, can be utilized as the extrinsic resource to aid the robot in performing object manipulation. Furthermore, by using an extrinsic supporting plane, grippers with simple structures can accomplish complex and challenging manipulation tasks. In Fig. \ref{f0}, we present a specific example to elucidate the concept of extrinsic manipulation using a plane. If the robot directly flips the T-nut through a single pick-and-place step, the manipulation will fail.
This is because, in certain scenarios (e.g., the initial scene depicted in the blue border), the inverse kinematics calculations of the robot are constrained by the environment, resulting in the robot approaching the object in awkward postures and even colliding with the plane (e.g., the scenes indicated in the red border). However, object flipping can be achieved if the robot relies on a plane to adjust the relative position between the object and the gripper.
By utilizing the object’s placements afforded by the extrinsic plane, the robot achieves the flipping of the object through regrasping it, while maintaining feasible and collision-free postures (e.g., the scenes indicated in the green border). Concretely, the intermediate stable placements of the object enable the robot to execute grasp configurations that were previously unattainable in the object’s initial position. 

The key to achieving extrinsic manipulation through regrasping is to predict diverse placements of the object on a plane and then establish transformations between these placements with shared grasp configurations. Thus, extrinsic manipulation can be achieved by facilitating sequential pick-and-place operations. 
Many mesh-based methods \cite{wan2019preparatory,wan2016achieving} analyze the geometry of rigid objects to obtain placements on a plane, assuming that the object models are given. 
\IEEEpubidadjcol
However, the mesh models of objects are unknown in many manipulation scenarios. For example, assembly automation \cite{suarez2018can,shi2023sim} requires the robots to manipulate parts that have different geometries with their mesh models unknown.
Therefore, it is crucial to propose approaches that can generalize to unseen objects for accomplishing manipulation tasks in different scenarios.


Deep learning methods \cite{jiang2021novel,duan2022learning} have showcased promising generalization capability in vision-based manipulation tasks. In recent work, data-driven methods \cite{cheng2021learning,xu2022learning} have been proposed to predict the placements of objects on supporting items (e.g., boxes and containers), using point clouds. Although these methods can generalize to predict placements of novel objects on novel supports, their results occasionally suffer from object penetration. Additionally, these methods are presently suitable for objects with elongated geometry (e.g., wrenches and spoons). 
Nevertheless, relying on extrinsic supporting objects for placements may not be feasible or suitable for many objects. Firstly, matching appropriate supporting items for specific types of objects is a complex task. Secondly, the geometry of objects can result in diverse placements on the plane. For example, manipulating the T-nut shown in Fig. \ref{f0} is more appropriate on a plane rather than placing it on a container. 
Simeonov \textit{et al.} \cite{simeonov2020long} propose employing a deep neural network sampler to jointly sample the object placements and the grasp configurations of the gripper for achieving long-term object manipulation. However, this approach assumes that an action skeleton is predetermined and relies on multiple action primitives to accomplish the manipulation task.

The enumeration of object placements on a plane, classified by orientations, remains a challenging task in the context of extrinsic manipulation. We propose a data-driven approach that utilizes point clouds as input to predict diverse placements of the object on a plane. Our framework comprises neural network models that are carefully designed to address two essential aspects: stability and diversity of the predicted placements.
Our primary objective is to predict a diverse set of stable placements that allows for feasible grasp configurations in extrinsic manipulation. 
To tackle the challenge of long-horizon regrasping, Xu \textit{et al.} \cite{xu2022learning} have developed an algorithm that utilizes point clouds to construct manipulation graphs, where the placements serve as nodes, and the shared grasp configurations act as edges connecting placements. In line with this method, we calculate the grasp configurations to facilitate the linkage between the predicted placements of objects on a plane, enabling sequential pick-and-place operations to regrasp objects to target locations.

We conduct extensive experiments to evaluate our approach. The experimental results demonstrate that our approach outperforms the state-of-the-art approaches in terms of stability and diversity of predicted placements.
The main contributions of our work are as follows: 
(1) This work pioneers the use of neural network models in the field of enumerating precise and diverse object placements on a plane.
(2) We construct a large-scale dataset that encompasses placements of objects with diverse geometric characteristics and captures the contact information between objects and the plane.
(3) Our approach showcases its ability to predict precise and diverse stable placements for accomplishing extrinsic manipulation on a support plane.

\section{Related Work}

\subsection{Object Placements Prediction}

Object placements refer to the poses of objects where the objects can be placed on the supporting items. Object placements are crucial for extrinsic manipulation by regrasping. Previous research has explored different approaches for object placement prediction. Some pioneer works, such as Wan \textit{et al.} \cite{wan2019regrasp,wan2019preparatory}, use the convex hull of object models to determine facets for stable placements. Ma \textit{et al.} \cite{ma2018regrasp} tackle placements prediction on complex supports with mesh-based simulations. However, these methods face limitations when dealing with unseen objects in real-world scenarios, as complete object models may not be available. 
Recent advancements in deep learning models for visual perception understanding have led to the development of placement prediction methods based on visual perception. For instance,
ReorientBot \cite{wada2022reorientbot} generates object motions using learned selection models that are trained for visual scene understanding. Paxton \textit{et al.} \cite{paxton2022predicting} use a Mixture Density Network (MDN) to construct a Gaussian mixture model, which captures the probability distribution of specific placements. Cheng \textit{et al.} \cite{cheng2021learning} introduce a two-stage pipeline for obtaining placements on supporting objects. Notably, they train their generation model with the supervision of transformed point clouds, whereas neural network models in \cite{xu2022learning} are trained with the supervision of 6-DoF placement poses. 
In our work, we train our generation model with carefully designed objectives, focusing solely on the correlative orientations of placements.

\subsection{Analysis of Contact between Objects}
There is a considerable body of research focusing on the study of contact scenarios between objects, particularly in terms of stability.
Hou \textit{et al.} \cite{hou2019reorienting} conduct a mechanical analysis of given meshes and develop a pivoting approach that enables objects to be reoriented while maintaining contact with the table plane.   
More recent works have explored employing neural network models trained with manually defined labels to estimate placement poses where contact occurs. 
OmniHang \cite{you2021omnihang} introduces two stages to predict and match contact points on two objects for stable hanging. Similarly, \cite{cheng2021learning} learns to predict offsets for separate point sets of a point cloud, allowing them to identify key contact points on the object. 
Upright-Net \cite{pang2022upright} infers upright placements of objects by extracting points that make contact with the supporting plane. 
Based on the slippage analysis theory, StablePose \cite{shi2021stablepose} uses geometrically stable patches (such as planes or cylinders) extracted from point clouds to enhance object pose inference. 
In contrast to predicting contact points, some studies concentrate on regressing the object's pose in contact with supporting items. For instance, Newbury \textit{et al.} \cite{newbury2021learning} use a Convolutional Neural Network (CNN) to infer the rotation of upright placement from depth images, while another CNN evaluates the quality of the resulting placement after rotation.  
In our work, we utilize neural network models to generate the auxiliary plane that makes contact with objects in non-canonical poses. We formulate the prediction of contact poses for objects as a pose regression problem of the supporting plane.

\subsection{Placement Classification}

To extract feature embeddings from point clouds for placement classification, prior works have resorted to PointNet++ networks introduced by Qi \textit{et al.} \cite{qi2017pointnet++}. For instance, Paxton \textit{et al.} \cite{paxton2022predicting} use a single PointNet++ network to extract features from input point clouds for realistic placement discrimination. However, a single PointNet++ network faces challenges in discriminating between stable placements and complex unstable placements, particularly those involving contact with the supporting plane. To mitigate this limitation, \cite{cheng2021learning} proposes a joint training approach for placement classification and contact point regression. Xu \textit{et al.} \cite{xu2022learning} use a PointNet++ network supervised by different types of unstable placements collected in the simulation to classify different types of placements, which is formulated as a multiple classification problem. More recently, Jiang \textit{et al.} \cite{jiang2021synergies} demonstrate that the implicit representation method enhances capturing the continuous feature for joint learning of multiple tasks. Motivated by this, we employ the feature manipulation techniques of both PointNet++ and the implicit representation method for discriminating between different placements.

\section{Approach}

\begin{figure*}[t]
\centering
\includegraphics[width=\linewidth]{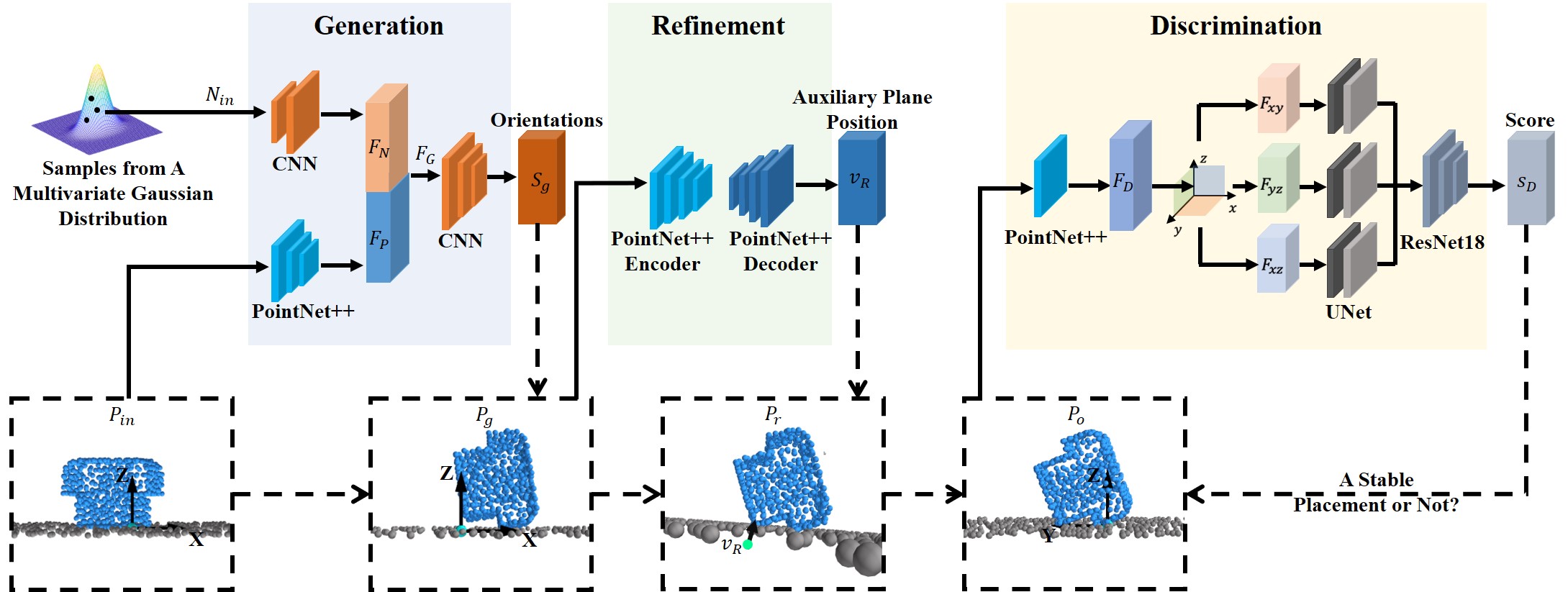}
\caption{Our three-stage framework for placement prediction. The solid lines represent the inputs of the models, the process of feature mapping, and the outputs of the models. The dashed lines represent point clouds and the process of transformation. }
\label{pic_pipline}
\end{figure*}

The objective of our proposed approach is to predict precise and diverse placements on the plane using point clouds. These placements, along with computed grasp configurations that connect these placements, are used to facilitate extrinsic manipulation of objects. As depicted in Fig. \ref{pic_pipline}, we propose a three-stage framework consisting of an orientation generation stage (Sec. \ref{3-1}), a placement refinement stage (Sec. \ref{3-2}), and a binary classification stage (Sec. \ref{sec3.3}) on refined placements. To enhance the diversity and stability of the predicted placements, each stage incorporates specific strategies. 
In the first stage, the orientations of the placements are generated through a neural network model. This method is adopted due to the pivotal role of orientations in discerning different placement types that are explained in Sec. \ref{sec4}.
Subsequently, in the second stage, the relative positions between the extrinsic supporting plane and the objects transformed by the generated orientations are refined. Finally, the third stage employs a classifier model to differentiate the refined placements as either stable or unstable and score them. 

The point clouds passed to the framework are obtained by applying the technique described in \cite{breyer2021volumetric} to capture depth images from multiple viewpoints around the object and then fuse them to generate a perceived point cloud. This operation aims to acquire as many points as possible for the objects. Importantly, the order of points remains preserved throughout the entire framework. 
The predicted placements categorized into different types are used to compute shared grasp configurations between them. 
With these grasp configurations, the robot can sequentially pick and place the object in extrinsic manipulation relying on the plane to regrasp the object to the goal position.

\subsection{Orientation Generation}
\label{3-1}



The primary objective of the first stage is to leverage a neural network model to generate different orientations from random noises based on different input point clouds (i.e., different observed objects). To achieve this, we follow the work \cite{xu2022learning} to adopt a network model of encoder-decoder architecture. In this stage, point clouds and random noises from a multivariate Gaussian distribution are jointly used as inputs. The inclusion of random noises in the input serves the purpose of preventing the generated orientations from converging to a single orientation \cite{goodfellow2020generative}. The PointNet++ network is employed to encode point clouds, motivated by its wide adoption in different applications \cite{cheng2021learning,xu2022learning,paxton2022predicting}. Specifically, a PointNet++ network with set abstraction layers encodes the point cloud $P_{in}$ into a feature map $F_{P}$. A convolutional neural network (CNN) encodes $K$ random noises $N_{in}$ into a feature map $F_{N}$. Subsequently, $F_{P}$ and $F_{N}$ are concatenated into a feature map $F_{G}$ and then fed into a CNN to output orientations $S_g$. 

We introduce a novel loss function to supervise the training. 
Specifically, in order to compare the generated orientations with the ground truths, we enhance the Chamfer distance \cite{fan2017point} by substituting the Euclidean distance with the Geodesic distance \cite{zhou2019continuity}. The Geodesic distance provides a more accurate representation of the angular distance between two orientations and is formulated as follows: 
\begin{equation}
d_{geo}(R_g, R_T) = arccos(\frac{tr(R_g R_T^{-1})-1}{2}),
\label{equ_geo}
\end{equation}
where $R_g$ is a generated rotation matrix and $R_T$ is a ground-truth rotation matrix. The result of $R_g R_T^{-1}$ is still a rotation matrix, and the trace of rotation matrix\cite{sarabandi2019survey} lies within the range of $[-1,3]$. Consequently, we have $tr(R_g R_T^{-1}) \in [-1,3]$. However, the function $d_{geo}(R_g, R_T)$ is not differentiable at zero when $tr(R_g R_T^{-1}) = 3$, which can pose challenges during the back-propagation in training. To overcome this issue, we modify the original expression of $d_{geo}(R_g, R_T)$ by representing it as a polynomial.
In order to strike a balance between fitting accuracy and computational complexity during training, we employ a tenth-order polynomial $\tilde{d}_{geo}(R_g, R_T)$ to approximate $d_{geo}(R_g, R_T)$. The polynomial is defined as follows:
\begin{equation}
   \tilde{d}_{geo}(R_g, R_T) = (tr(R_g R_T^{-1})-3)\sum_{i=0}^{9} a_itr(R_g R_T^{-1})^i.
\end{equation}
Same as ${d}_{geo}(R_g, R_T)$, the polynomial $\tilde{d}_{geo}(R_g, R_T)$ also satisfies the condition that the result is zero when $tr(R_g R_T^{-1})=3$. 
Ultimately, the loss function is expressed as:
\begin{equation}
\begin{split}
        \mathcal{L}_{geo} &= \sum_{R_g \in S_g} \min _{R_T \in S_T}\tilde{d}_{geo}(R_g, R_T)\\
        &+\sum_{R_T \in S_T} \min _{R_g \in S_g}\tilde{d}_{geo}(R_g, R_T),
\end{split}
\label{equ_loss}
\end{equation}
where $S_g$ and $S_T$ represent the generated and ground-truth orientations, respectively.  
Given a generated orientation, the input $P_{in}$ is transformed to a rough placement $P_g$ as shown in Fig. \ref{pic_pipline}.

\subsection{Placement Refinement} 
\label{3-2}


In the second stage, a PointNet++ network is employed to adjust the relative position between the object’s point cloud transformed by a generated orientation and the extrinsic supporting plane. Our method involves generating an auxiliary plane relative to the transformed point cloud of the object in the world coordinate frame. Subsequently, the generated auxiliary plane is aligned with the extrinsic supporting plane, ensuring that the relative position between the generated auxiliary plane and the transformed point cloud of the object remains unchanged. Then, the refined placement of the object is obtained by following these procedures.
In Fig. \ref{pic_pipline}, the generated auxiliary plane is represented by a vector denoted as $v_R$, which possesses a specific length and is perpendicular to the generated auxiliary plane. Mathematically, the generated auxiliary plane can be expressed as:
\begin{equation}
   ax + by + cz = a^2 + b^2 + c^2,
\end{equation} 
where $v_R=(a, b, c)$ is a vector defined in the world coordinate frame.

We employ a PoinNet++ network with set abstraction layers and feature propagation layers to process the point cloud $P_g$ comprising $M$ points to obtain the per-point feature vector for each point $p_i \in P_g $. 
This network is subsequently followed by fully connected layers with ReLU activation and batch normalization.
Consequently, each per-point feature vector, denoted as $v^d_i$, is reduced to a dimension of 3.  
To clarify the interpretation of a per-point feature vector $v^d_i$, we define each per-point feature vector as the subtraction result between the generated auxiliary plane vector $v_R$ and a point vector $p_i$ (i.e., $v^d_i = v_R - p_i$). This definition enables us to capture the spatial relationship between the vector of input points and the generated auxiliary plane vector.

Considering diverse transformed point clouds of an object as input of the second stage, the loss function is designed to infer distinct auxiliary planes for the same object transformed by different orientations. In contrast to the loss function in the previous work \cite{cheng2021learning}, which predicts a target point using points within a specific region of the object, our loss function establishes the correlation between each point in the input point cloud and the auxiliary plane vector. By incorporating a greater number of points into the training objective, the accuracy of the inferred auxiliary plane vectors can be improved. Consequently, we define the training objective as follows:
\begin{equation}
\begin{split}
   \mathcal{L}_{refine} &= \mathcal{L}_{field} + \mathcal{L}_{var}\\
   &=\alpha \frac{\sum_{i = 1}^{M} smoothl1loss( v_{GT} - p_i, v^d_i )}{M} \\
   &+ \beta \frac{\sum_{i = 1}^{M} (p_i + v^d_i - \frac{\sum_{i = 1}^{M} p_i + v^d_i }{M})^2}{M},
\end{split}
\end{equation} 
where $\mathcal{L}_{field}$, $\mathcal{L}_{var}$ are mentioned in \cite{cheng2021learning} as the field loss and the variance loss, and $\alpha$, $\beta$ are weights. The ground-truth auxiliary plane vector is denoted as $v_{GT}$. $smoothl1loss$ is smooth L1 loss. 

\begin{figure}[h]
      \centering
      \includegraphics[width=\linewidth]{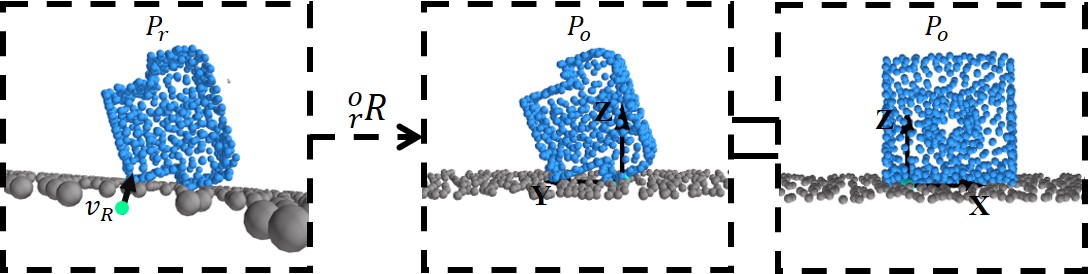}
      \caption{$P_{r}$ is transformed by ${}_r^{o}R$. The two images on the right side are $P_{o}$ at different view angles.}
      
      \label{trans_B}
\end{figure}

After obtaining the generated vector $v_R=\frac{\sum_{i = 1}^{M} p_i + v^d_i }{M}$, we obtain the point cloud $P_r$ as shown in Fig. \ref{pic_pipline}. The object's point cloud remains unchanged, while the point cloud of the auxiliary plane is positioned perpendicular to the generated vector $v_R$. 
To ensure alignment between the auxiliary plane and the extrinsic supporting plane in the world coordinate frame, $P_r$ is transformed to $P_{o}$. This transformation is achieved by applying a rotation matrix ${}_r^{o}R$ to the point cloud $P_r$, given by:
\begin{equation}
  P_{o} = {}_r^{o}R \cdot (P_r - v^{\rm T} \cdot v_R),
\end{equation}
where ${}_r^{o}R$ is obtained by matching the auxiliary plane and the extrinsic supporting plane through the shortest path \cite{cid2018lipschitz}. $v^{\rm T}$ represents a vector of ones, and its purpose is to extend the dimension, such that each point in $P_r$ is subtracted by $v_R$. 
Fig. \ref{trans_B} shows the transformation process.
During this transformation, the relative position between the object and the auxiliary plane remains unchanged, while the placement is refined.

\subsection{Placement Discrimination}
\label{sec3.3}

Discriminating between stable and unstable placements is still challenging, particularly due to the various interaction cases between objects and subtle differences between placements. For example, the distinction between a stable placement where the object makes contact with the plane, and an unstable placement where the object also makes contact but lacks stability can be minimal. Previous studies \cite{cheng2021learning,xu2022learning,paxton2022predicting} have utilized PointNet++ networks with multiple set abstraction layers to identify stable placements of objects. However, while a single PointNet++ network performs well in classifying point clouds with varying object geometries \cite{qi2017pointnet++}, it falls short in distinguishing point clouds of the same object with different poses. Based on the PointNet++ network, we integrate the feature manipulation techniques introduced in GIGA \cite{jiang2021synergies} to design a classifier for classifying stable and unstable placements using point clouds.

Our discriminator firstly uses a PointNet++ network with one set abstraction layer to extract a feature embedding $F_D$ of the input point cloud $P_o$. Subsequently, we construct orthogonal feature planes ($F_{xy}$, $F_{yz}$, and $F_{xz}$) by vertically projecting the feature embedding $F_D$ onto the pixel cells of the respective feature planes. Each feature plane encompasses pixel-wise features within the feature space. Subsequently, U-Net \cite{ronneberger2015u} networks process these feature planes to perform feature inpainting, as introduced in \cite{jiang2021synergies}. The resulting features are then concatenated and passed to a ResNet18 \cite{he2016deep} network, which maps the concatenated feature to a 2D output of predicates. The score $s_{D}$ for each input point cloud is obtained based on this 2D output.
During training, the cross-entropy loss \cite{xu2022learning} is adopted as the training objective, with binary classification supervised by the ground-truth labels presented in Sec. \ref{sec4}. To guarantee accurate and diverse placements are used to calculate grasp configurations, input placements with output scores $s_{D}$ above 0.92 are considered and sorted into different types based on their orientations using our clustering approach described in Sec. \ref{sec4}.


\subsection{Grasp Configuration Calculation}
\label{sec3.4}

\begin{figure}[h]
      \centering
      \includegraphics[width=\linewidth]{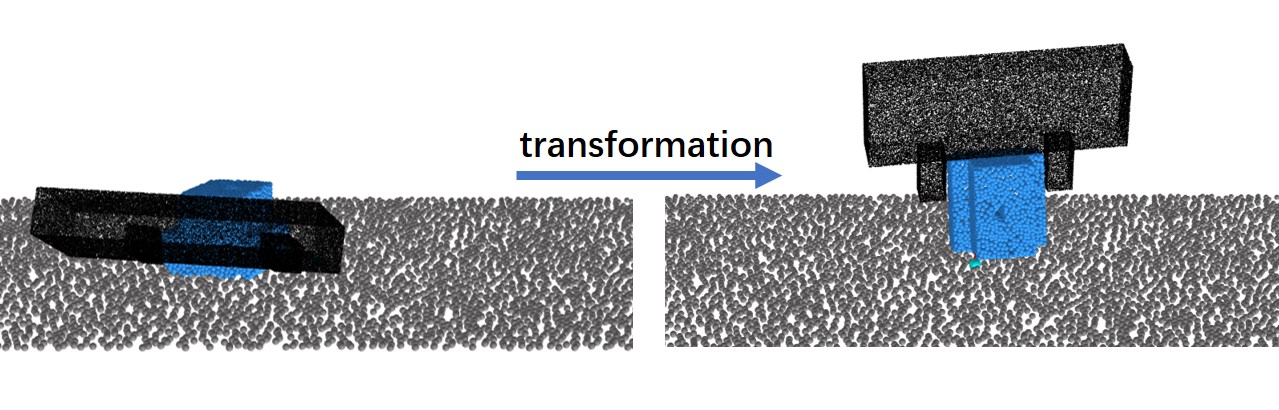}
      \caption{An example of a shared grasp configuration.}
      
      \label{grasp-conf}
\end{figure}

The use of predicted placements on the plane is crucial for regrasping objects in the context of extrinsic manipulation. Regrasping the object to the goal position is achieved by employing pick-and-place primitives, necessitating the calculation of grasp configurations for grasping and placing objects. 
The shared grasp configurations between the predicted placements are computed with a method proposed in \cite{xu2022learning}. The calculation process involves determining the grasp points of the gripper and sampling feasible approaching directions of the gripper that also remain viable when the placements are transformed by predicted transformations. Additionally, the feasible inverse kinematics of the robot arm is taken into account for the associated approaching direction of the gripper. An example in Fig. \ref{grasp-conf} is provided to demonstrate a feasible shared grasp configuration between two placements. 

\begin{figure*}[t]
      \centering
      \includegraphics[width=\linewidth]{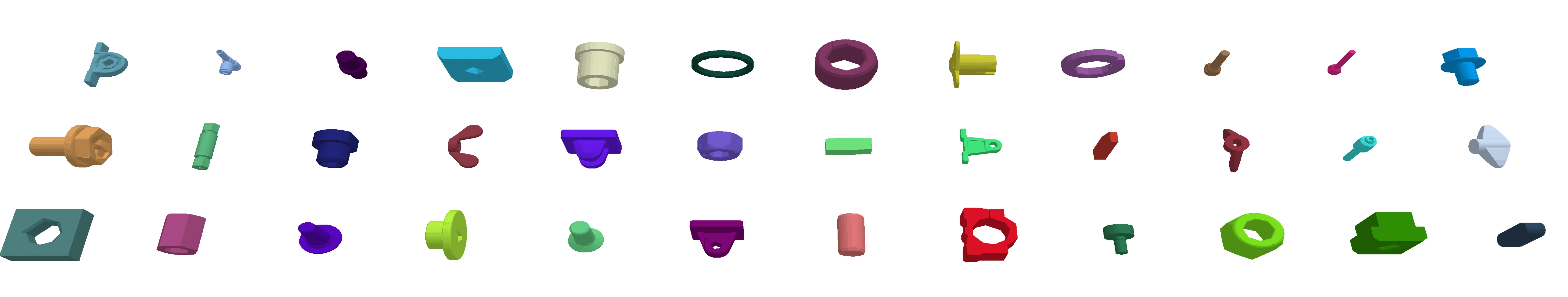}
      
      \caption{Models of part of the objects in our dataset. Objects in the first two rows are in the training set. Objects in the last row are in the test set and used for experiments.}
      
      \label{pic_dataset_total}
\end{figure*}

\begin{figure*}[t]
      \centering
      \includegraphics[width=\linewidth]{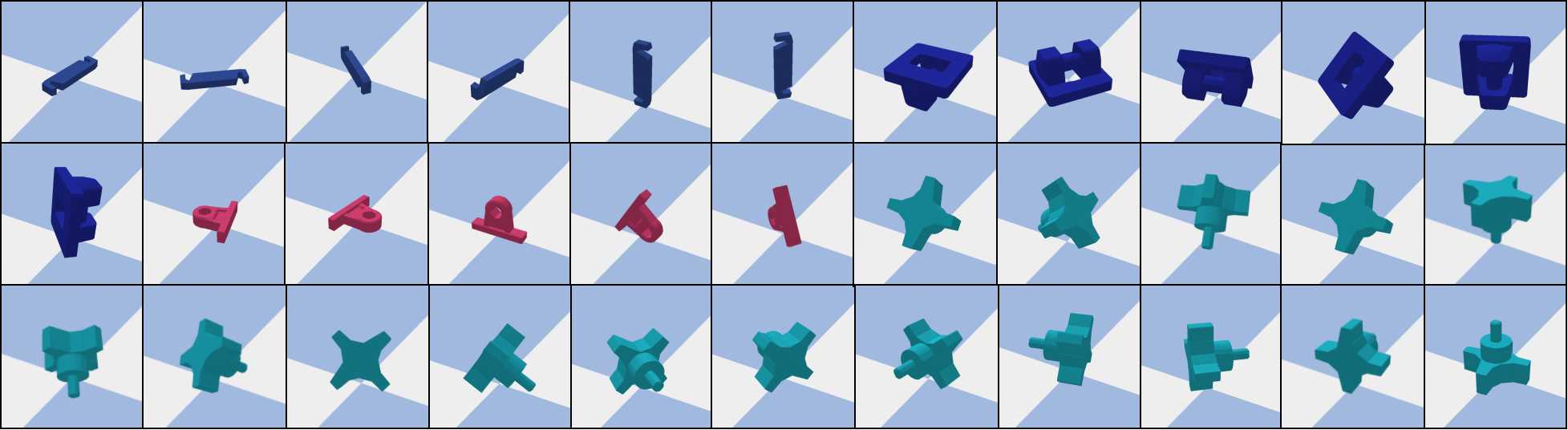}
      \caption{Collected stable placements of several objects in the training set. The stable placements are classified according to the orientations.}
      \label{pic_dataset_single}
\end{figure*}

\section{Dataset And Training}
\label{sec4}

\subsection{Dataset}
There is currently no available dataset specifically designed for diverse object placements on a plane. To address this gap, we construct a dataset of object placements utilizing the mesh models from the Mechanical Components Benchmark (MCB) \cite{sangpil2020large}. The MCB consists of 3D objects with diverse geometric shapes. Our dataset encompasses a total of 374 objects across 25 categories. The training set includes 275 objects, and the test set contains 99 objects with shapes that differ from those in the training set. As depicted in Fig. \ref{pic_dataset_total}, we showcase a subset of the objects in our dataset to demonstrate the range of shape variations.

To obtain the ground-truth object placements, we utilize PyBullet \cite{coumans2016pybullet} to perform random dropping simulations. Initially, we load objects from our dataset into PyBullet with random positions above the extrinsic supporting plane. Subsequently, each simulation process runs for a minimum of 10 s. Each simulation is halted once the object's velocity drops below a threshold, ensuring that the object settles into a stable placement under gravity. Finally, we record the translation and orientation of a stable placement. To obtain diverse placements, each object is subjected to 1000 simulations in PyBullet. Consequently, we acquire a total of 374,000 ground-truth placements of 374 objects in the world coordinate frame.

Moreover, we utilize the continuous representation of orientations \cite{zhou2019continuity} and apply a clustering approach, incorporating MeanShift \cite{1055330}, to categorize the collected placements into distinct types. The number of clusters can be determined automatically based on the distribution of orientations of the collected placements.
We assume that performing rotations around the $z$ axis and translations along the $xy$ plane on a classified placement in the world coordinate frame will yield placements of the same type as the original classified placement.
As depicted in Fig. \ref{pic_dataset_single}, the placements of objects in the training set are classified into their respective types.
The numbers of different placement types for each object are carefully balanced in the training data of the first stage. The orientations of the classified placements are used as supervision.



In addition to recording stable placements, we also record the contact points on the object in the stable placement state as well as the unstable pose of the object in mid-air during each simulation. Specifically, we select three non-collinear contact points from each placement, which are used to describe the position of the auxiliary plane. By establishing the transformation between a stable placement and an unstable pose, we are able to obtain the coordinates of the three points corresponding to the unstable pose. These coordinates allow us to calculate a vector that represents the auxiliary plane in the world coordinate frame. The unstable poses of an object, along with their corresponding auxiliary planes, are utilized for training the placement refinement model.
Furthermore, we collect unstable placements that make contact with the extrinsic supporting plane to train our discriminator.

\subsection{Training Details}
To implement the neural network models for all three stages, we utilize the PyTorch \cite{paszke2019pytorch} framework. The training process is performed on an NVIDIA GeForce GTX 3090 GPU. We employ the Adam optimizer for training. All training loss curves and validation curves converge.



\section{Experiments}

We conduct a series of experiments to validate the effectiveness of our framework in predicting diverse stable placements for novel objects with varying geometries. Specifically, the performance of the orientation generation, position refinement, and placement discrimination stages are evaluated through comparative analyses and self-ablation experiments. To further validate the practical applicability of our method, simulation and real-robot experiments to manipulate objects are also conducted.

\subsection{Simulation Experiments}


\begin{table*}[hb] 
   \caption{The Accuracy of Predicted Placements}
   \label{table_acc}
   \centering
   \LARGE
   
   \resizebox{\linewidth}{!}{
   \begin{tabular}{lcccccccccccccccc}
   \toprule
   \multicolumn{1}{c}{Approaches}
   & \multicolumn{1}{c}{\begin{tabular}[c]{@{}c@{}}Square\\ Nut\end{tabular}}  
   & \multicolumn{1}{c}{\begin{tabular}[c]{@{}c@{}}Standard\\ Fitting\end{tabular}}  
   & \multicolumn{1}{c}{\begin{tabular}[c]{@{}c@{}}Conventional\\ Rivet\end{tabular}}  
   & Collar  
   & \multicolumn{1}{c}{\begin{tabular}[c]{@{}c@{}}Tapping\\ Screw\end{tabular}}  
   & \multicolumn{1}{c}{\begin{tabular}[c]{@{}c@{}}Flanged Block \\ Bearing\end{tabular}}  
   & \multicolumn{1}{c}{\begin{tabular}[c]{@{}c@{}}Cylindrical\\ Pin\end{tabular}}  
   & Clamp  
   & Bush  
   & Locknut  
   & T-nut  
   & Key  
   & \textbf{Average}\\
   \midrule
   

   $\mathrm{G_{L2G\text{\cite{cheng2021learning}}}+R_{ours}+D_{ours}}$
&\textbf{1.000} &{0.800} &{0.682} &\textbf{1.000} &\textbf{0.857} &{0.857} &{0.686} &{0.929} &{0.579} &\textbf{0.952} &{0.786} &{0.786} &{0.826} \\
   $\mathrm{G_{CD\text{\cite{fan2017point}}}+R_{ours}+D_{ours}}$
   &\textbf{1.000} &{0.477} &{0.773} &{0.600} &{0.621} &{0.382} &\textbf{0.900} &{0.961} &{0.630} &{0.810} &{0.971} &{0.907} &{0.753} \\

   $\mathrm{G_{random}+R_{ours}+D_{ours}}$
   &\textbf{1.000} &\textbf{0.857} &{0.759} &{0.955} &\textbf{0.857} &{0.857} &{0.571} &\textbf{1.000} &\textbf{1.000} &{0.714} &\textbf{1.000} &{0.857} &{0.869} \\
   $\mathrm{G_{ours}+R_{ours}+D_{ours}}$
   &\textbf{1.000} &\textbf{0.857} &\textbf{0.857} &\textbf{1.000} &{0.786} &\textbf{1.000} &{0.875} &\textbf{1.000} &{0.571} &{0.905} &\textbf{1.000} &\textbf{1.000} &\textbf{0.904} \\
   \midrule
   
   $\mathrm{G_{ours}+D_{ours}}$
   &{0.619} &{0.400} &{0.692} &{0.533} &{0.609} &{0.429} &\textbf{1.000} &{0.700} &{0.381} &{0.458} &{0.643} &{0.429} &{0.574} \\
   $\mathrm{G_{ours}+R_{ours}+D_{ours}}$
   &\textbf{1.000} &\textbf{0.857} &\textbf{0.857} &\textbf{1.000} &\textbf{0.786} &\textbf{1.000} &{0.875} &\textbf{1.000} &\textbf{0.571} &\textbf{0.905} &\textbf{1.000} &\textbf{1.000} &\textbf{0.904} \\
   \midrule

   $\mathrm{G_{ours}+R_{ours}+D_{PN2\text{\cite{qi2017pointnet++}}}}$
   &{0.857} &{0.286} &{0.571} &{0.500} &{0.714} &{0.571} &{0.429} &{0.429} &\textbf{1.000} &{0.714} &{0.524} &{0.143} &{0.562} \\

   $\mathrm{G_{ours}+R_{ours}+D_{random}}$
   &{0.300} &{0.350} &{0.350} &{0.050} &{0.250} &{0.150} &{0.250} &{0.350} &{0.400} &{0.200} &{0.250} &{0.100} &{0.250} \\
   
   $\mathrm{G_{ours}+R_{ours}+D_{ours}}$
   &\textbf{1.000} &\textbf{0.857} &\textbf{0.857} &\textbf{1.000} &\textbf{0.786} &\textbf{1.000} &\textbf{0.875} &\textbf{1.000} &{0.571} &\textbf{0.905} &\textbf{1.000} &\textbf{1.000} &\textbf{0.904} \\
   \midrule
   
   Baseline\cite{cheng2021learning}
    &\textbf{1.000} &\textbf{0.857} &{0.714} &{0.000} &{0.643} &{0.000} &{0.643} &\textbf{1.000} &\textbf{1.000} &\textbf{0.929} &{0.286} &{0.571} &{0.637} \\


   $\mathrm{G_{ours}+R_{ours}+D_{ours}}$
   &\textbf{1.000} &\textbf{0.857} &\textbf{0.857} &\textbf{1.000} &\textbf{0.786} &\textbf{1.000} &\textbf{0.875} &\textbf{1.000} &{0.571} &{0.905} &\textbf{1.000} &\textbf{1.000} &\textbf{0.904} \\
   \bottomrule
   \end{tabular}}
   \raggedright
   \end{table*}


\begin{table*}[hb]

   \caption{The Diversity of Predicted Placements}
   \label{table_div}
   \centering
   \LARGE
   
   \resizebox{\linewidth}{!}{
   \begin{tabular}{lcccccccccccccccc}
   \toprule
   \multicolumn{1}{c}{Approaches}
   & \multicolumn{1}{c}{\begin{tabular}[c]{@{}c@{}}Square\\ Nut\end{tabular}}  
   & \multicolumn{1}{c}{\begin{tabular}[c]{@{}c@{}}Standard\\ Fitting\end{tabular}}  
   & \multicolumn{1}{c}{\begin{tabular}[c]{@{}c@{}}Conventional\\ Rivet\end{tabular}}  
   & Collar  
   & \multicolumn{1}{c}{\begin{tabular}[c]{@{}c@{}}Tapping\\ Screw\end{tabular}}  
   & \multicolumn{1}{c}{\begin{tabular}[c]{@{}c@{}}Flanged Block \\ Bearing\end{tabular}}  
   & \multicolumn{1}{c}{\begin{tabular}[c]{@{}c@{}}Cylindrical\\ Pin\end{tabular}}  
   & Clamp  
   & Bush  
   & Locknut  
   & T-nut  
   & Key  
   & \textbf{Average}\\


   \midrule
   $\mathrm{G_{L2G\text{\cite{cheng2021learning}}}+R_{ours}+D_{ours}}$
   &{0.000} &{0.000} &\textbf{1.000} &\textbf{0.857} &{0.000} &\textbf{1.000} &\textbf{1.000} &{0.333} &{0.000} &{0.250} &{0.444} &\textbf{1.000} &{0.490} \\
   $\mathrm{G_{CD\text{\cite{fan2017point}}}+R_{ours}+D_{ours}}$
   &\textbf{1.000} &{0.889} &\textbf{1.000} &{0.667} &\textbf{1.000} &{0.500} &\textbf{1.000} &{0.200} &\textbf{1.000} &{0.571} &{0.400} &\textbf{1.000} &{0.769} \\
   $\mathrm{G_{random}+R_{ours}+D_{ours}}$
   &{0.500} &\textbf{1.000} &\textbf{1.000} &{0.667} &\textbf{1.000} &{0.500} &\textbf{1.000} &\textbf{0.400} &\textbf{1.000} &{0.857} &{0.400} &\textbf{1.000} &{0.777} \\
   $\mathrm{G_{ours}+R_{ours}+D_{ours}}$
   &{0.500} &{0.889} &\textbf{1.000} &{0.667} &\textbf{1.000} &{0.500} &\textbf{1.000} &\textbf{0.400} &\textbf{1.000} &\textbf{1.000} &\textbf{0.800} &\textbf{1.000} &\textbf{0.813} \\
   \midrule

   $\mathrm{G_{ours}+D_{ours}}$
   &\textbf{1.000} &\textbf{0.889} &\textbf{1.000} &\textbf{1.000} &\textbf{1.000} &{0.250} &{0.500} &\textbf{0.800} &\textbf{1.000} &\textbf{1.000} &{0.400} &{0.667} &{0.792} \\
   $\mathrm{G_{ours}+R_{ours}+D_{ours}}$
   &{0.500} &\textbf{0.889} &\textbf{1.000} &{0.667} &\textbf{1.000} &\textbf{0.500} &\textbf{1.000} &{0.400} &\textbf{1.000} &\textbf{1.000} &\textbf{0.800} &\textbf{1.000} &\textbf{0.813} \\
   \midrule
   
   $\mathrm{G_{ours}+R_{ours}+D_{PN2\text{\cite{qi2017pointnet++}}}}$
   &{0.000} &{0.889} &\textbf{1.000} &\textbf{1.000} &\textbf{1.000} &\textbf{0.750} &\textbf{1.000} &{0.200} &\textbf{1.000} &\textbf{1.000} &\textbf{0.800} &{0.333} &{0.748} \\
   $\mathrm{G_{ours}+R_{ours}+D_{random}}$
   &{0.000} &\textbf{1.000} &\textbf{1.000} &{0.667} &\textbf{1.000} &{0.500} &\textbf{1.000} &\textbf{0.600} &\textbf{1.000} &\textbf{1.000} &{0.600} &{0.333} &{0.725} \\
   $\mathrm{G_{ours}+R_{ours}+D_{ours}}$
   &\textbf{0.500} &{0.889} &\textbf{1.000} &{0.667} &\textbf{1.000} &{0.500} &\textbf{1.000} &{0.400} &\textbf{1.000} &\textbf{1.000} &\textbf{0.800} &\textbf{1.000} &\textbf{0.813} \\

   \midrule
   Baseline\cite{cheng2021learning}
   &{0.000} &{0.000} &\textbf{1.000} &{0.571} &{0.000} &\textbf{1.000} &\textbf{1.000} &{0.000} &{0.000} &{0.000} &{0.000} &{0.500} &{0.339} \\

   $\mathrm{G_{ours}+R_{ours}+D_{ours}}$
   &\textbf{0.500} &\textbf{0.889} &\textbf{1.000} &\textbf{0.667} &\textbf{1.000} &{0.500} &\textbf{1.000} &\textbf{0.400} &\textbf{1.000} &\textbf{1.000} &\textbf{0.800} &\textbf{1.000} &\textbf{0.813} \\
   \bottomrule
   \end{tabular}}
   \raggedright
   \footnotesize{
      \\[2pt]
      The comparison results of generator performance, refinement performance, discriminator performance, and the baseline. In each part, the results of our approach are reported in the bottom row and the best results are shown in bold.}
   \end{table*}


We construct our simulated environment using PyBullet \cite{coumans2016pybullet}. 
As shown in Fig. \ref{pic_dataset_total}, the objects used in the simulation experiments are from the test set, ensuring that they are unseen during the training. 
The ground-truth labels for object placements in the test set, encompassing both stable placements and unstable positions, are introduced in Sec. \ref{sec4}. In contrast, the predicted placements of objects in the test set are generated using the approaches for comparative analysis. 
During the comparison of different approaches for a given object, the input point cloud remains constant, and the number of random noises provided as input is set to 1024. 
Subsequently, the test objects with the predicted placements are loaded into our simulated environment to record their final stabilized placements after the simulation.
The following metrics are used to evaluate different approaches:  
\begin{itemize}
\item Accuracy of predicted placements: This metric assesses the proportion of predicted placements that are verified as stable. We employ specific criteria to determine the stability of a predicted placement. Let $(O_{start}, h_{start})$ represent the orientation and height position of a predicted placement before simulation, and $(O_{end}, h_{end})$ denote the corresponding values after simulation.  We consider a predicted placement as stable if the following conditions are met:$\varDelta D = d_{geo}(O_{start}, O_{end})\times 180^\circ \leqslant 10^\circ $ and $\varDelta H = \left\lVert h_{start}-h_{end}\right\rVert \leqslant 2$ cm. 

\item Diversity of predicted placements: This metric quantifies the proportion of predicted placement types that have been verified as stable, in relation to the ground-truth placement types. Using the approach introduced in Sec. \ref{sec4}, we classify the predicted placements based on their orientations. To determine if a predicted type $O_{pred}$ corresponds to a ground-truth type $O_{gt}$, we establish a threshold $\varDelta d = d_{geo}(O_{pred}, O_{gt}) \times 180^\circ \leqslant 15^\circ $. Suppose a test object has $n$ types of stable placements. If an approach predicts $m$ types that are verified as stable and satisfy the threshold, the diversity of predicted placements can be calculated as $\frac{m}{n-1}$, where the initial placement type is excluded.
\end{itemize}

\subsubsection{Generation Performance}

To demonstrate the effectiveness of our orientation generation stage, we compare our approach ($G_{ours} +R_{ours} +D_{ours}$), where $G_{ours}$ denotes our orientation generation stage, $R_{ours}$ denotes our position refinement stage, and $D_{ours}$ denotes our placement discriminator, with the following variations of our approach:

\begin{itemize}
\item $G_{L2G}+R_{ours}+D_{ours}$:In this variation, we replace $G_{ours}$ in our approach with $G_{L2G}$ proposed in \cite{cheng2021learning}. The loss function in $G_{L2G}$ computes the Chamfer Distance \cite{fan2017point} between the point clouds transformed by predicted poses and the ground-truth point clouds. We retrain $G_{L2G}$ using our training set.

\item $G_{CD}+R_{ours}+D_{ours}$: Here, $G_{ours}$ in our approach is substituted with $G_{CD}$ introduced in \cite{xu2022learning}. The loss function in $G_{CD}$ computes the Chamfer Distance between the predicted orientations and the ground-truth orientations. We retrain $G_{CD}$ using the same training data as $G_{ours}$.

\item $G_{random}+R_{ours}+D_{ours}$: In $G_{random}$, the orientations in the first stage are randomly sampled. Specifically, we randomly sample Euler angles within the range of $[-\pi, \pi] $.

\end{itemize}

\begin{figure}[tp]
      \centering
      \includegraphics[width=\linewidth]{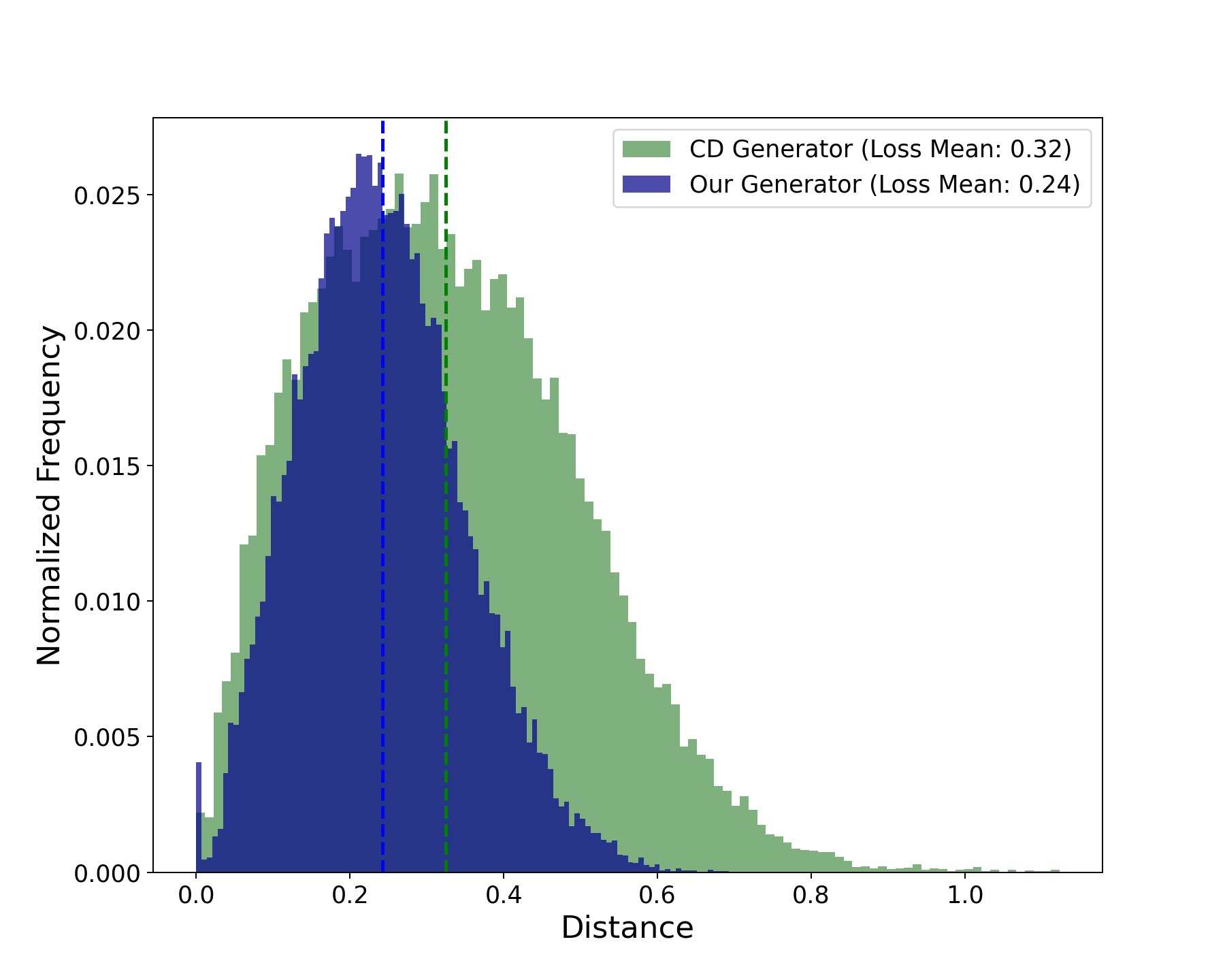}
      \caption{The distribution histogram of $d_{to}(R_T)$. CD Generator is $G_{CD}$ and the dotted lines are the means of different distributions.}
      \label{pic_exp_gen}
   \end{figure}

First, we assess the generalization ability of the learning-based models on novel objects from the test set. We train $G_{ours}$ and $G_{CD}$ for 500 epochs, and all models reach convergence. For each model, the distance between the generated orientations ($S_g$) and the ground-truth orientations ($S_T$) is calculated by $ d_{to}(R_T) =  \min\limits_{R_g \in S_g} \tilde{d}_{geo}(R_g, R_T)$, where $R_T \in S_T $. The resulting distribution histogram of $d_{to}(R_T)$ is visualized in Fig. \ref{pic_exp_gen}.  
The figure shows that the distance distribution of $G_{ours}$ is more tightly clustered around smaller distances compared to $G_{CD}$. Furthermore, $G_{ours}$ exhibits a higher proportion of small distance values than $G_{CD}$. 
Moreover, the mean distances for $G_{ours}$ and $G_{CD}$ are 0.24 and 0.32, respectively, indicating that the generated orientations from $G_{ours}$ are the closer approximation to the ground-truth orientations.


We further compare the performance of the aforementioned approaches on test objects from different categories.
As reported in the bottom rows of the first section in Tab. \ref{table_acc} and \ref{table_div}, our approach achieve the best performance compared to others in terms of the average accuracy and diversity of predicted placements, achieving 90.4\% accuracy and 81.3\% diversity. $R _{ours}$ and $D_{ours}$ remain constant across all approaches, which highlights the capability of $G_{ours}$ in generating accurate and diverse placements. 
For most objects listed in the tables, our approach surpasses or achieves comparable results to the best-performing alternative approaches in terms of accuracy and diversity of predicted placements.
These results can be attributed to the loss design in $G_{ours}$, which effectively aligns the generated orientations of the trained model with the ground-truth orientations. The advantage of utilizing $G_{ours}$, as opposed to $G_{random}$, is its ability to better fit the distribution of orientations through the learning process. 
In contrast, $G_{L2G}+R_{ours}+D_{ours}$ exhibits poor performance in diversity. Upon closer examination of its outputs, we observe that $G_{L2G}$ tends to generate limited types of placements for each object, resulting in a lack of diversity in its results.

\subsubsection{Refinement Performance}

We conduct self-ablation experiments to assess the effectiveness of the position refinement stage in our approach. The comparison results between our approach ($G_{ours} +R _{ours} +D_{ours}$) and the two-stage approach ($G_{ours} +D_{ours}$) are shown in the second part in Tab. \ref{table_acc} and \ref{table_div}.
As shown in Tab. \ref{table_acc}, our approach demonstrates higher accuracy in predicting placements for all objects except the cylindrical pin, compared to ($G_{ours} + D_{ours}$). 
Furthermore, our approach significantly improves the average accuracy of the predicted placements from 57.4\% in ($G_{ours} +D_{ours}$) to 90.4\%. These results indicate that $R _{ours}$ effectively refines the placements from the first stage, thereby transforming unstable placements into stable ones.
Tab. \ref{table_div} shows that the refinement stage contributes to enhancing the diversity of placements. Our approach outperforms ($G_{ours} + D_{ours}$) by 2.1\% in terms of the average diversity of predicted placements. 



\subsubsection{Discriminator Performance}

\begin{figure}[ht]
      \centering
      \includegraphics[width=8cm]{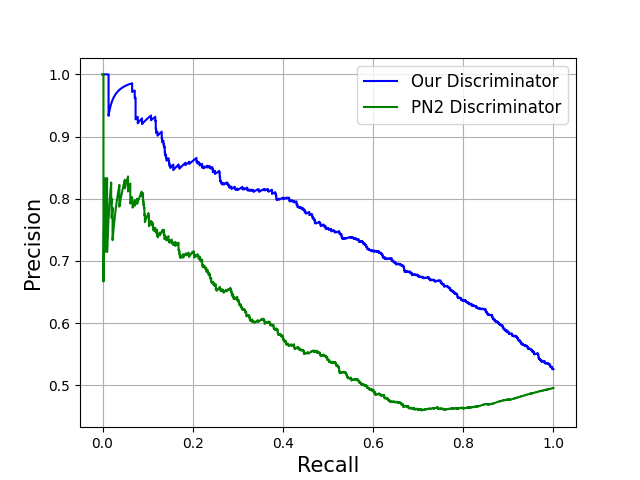}
      \caption{Precision-recall curves of different discriminators. PN2 discriminator is $D_{PN2}$.}
      \label{pic_dis_pr}
   \end{figure}


\begin{figure*}[t]
      \centering
      \includegraphics[width=\linewidth]{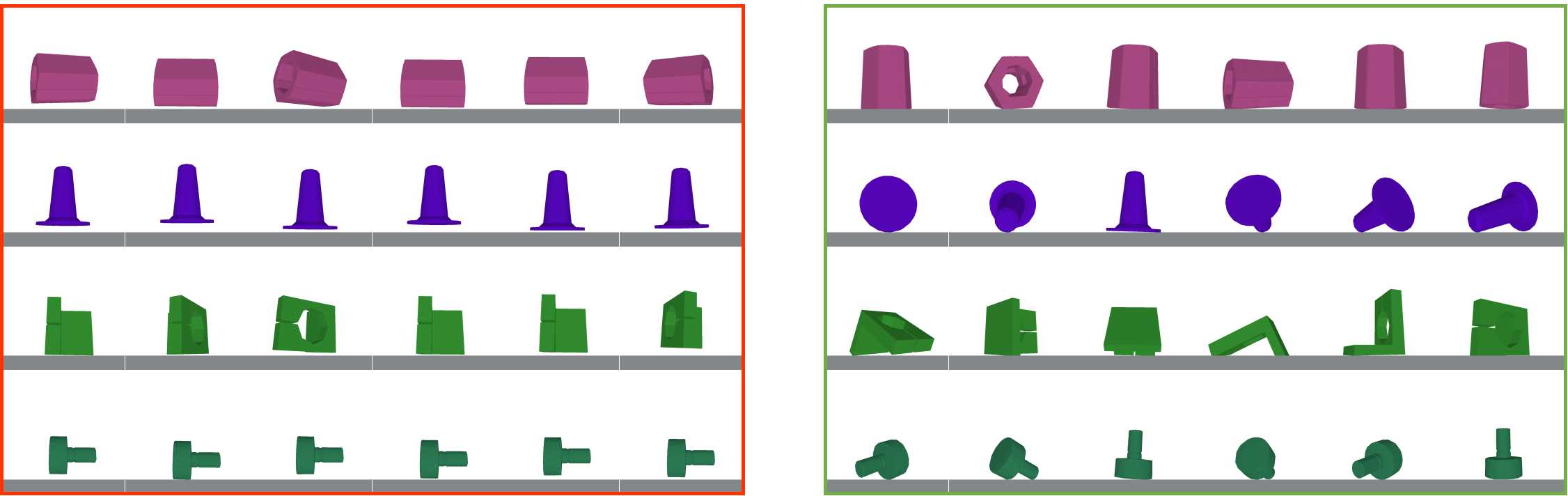}
      \caption{The predicted placements of the baseline (left, red border) and our approach (right, green border). These placements have been verified stable, and the figures are captured in PyBullet from a horizontal angle where the gray line represents the table plane.}
      \label{pic_exp_pred}
\end{figure*}


\begin{figure*}[hb]
\centering

\includegraphics[width=\linewidth,trim=0 0 0 160,clip]{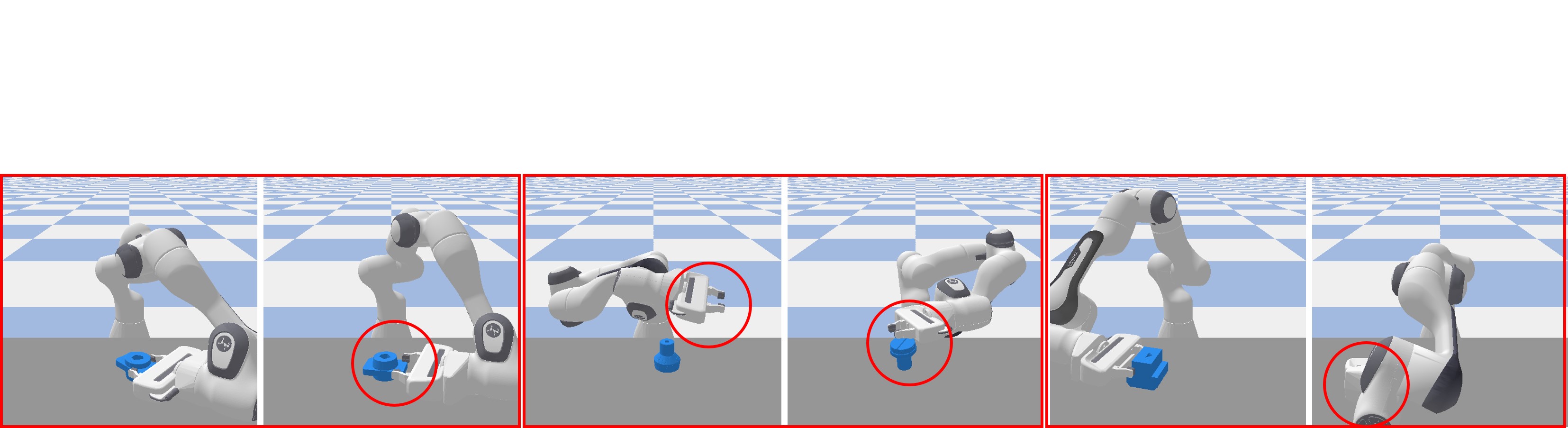}

\caption{Examples of constrained scenarios. The robot is unable to flip these objects with one pick-rotate-and-place step. Infeasible grasp configurations of the robot are marked with red circles. }

\label{makeup-exp0}
\end{figure*}

\begin{figure*}[htbp]
\centering

\includegraphics[width=0.99\linewidth,trim=0 30 0 150,clip,frame]{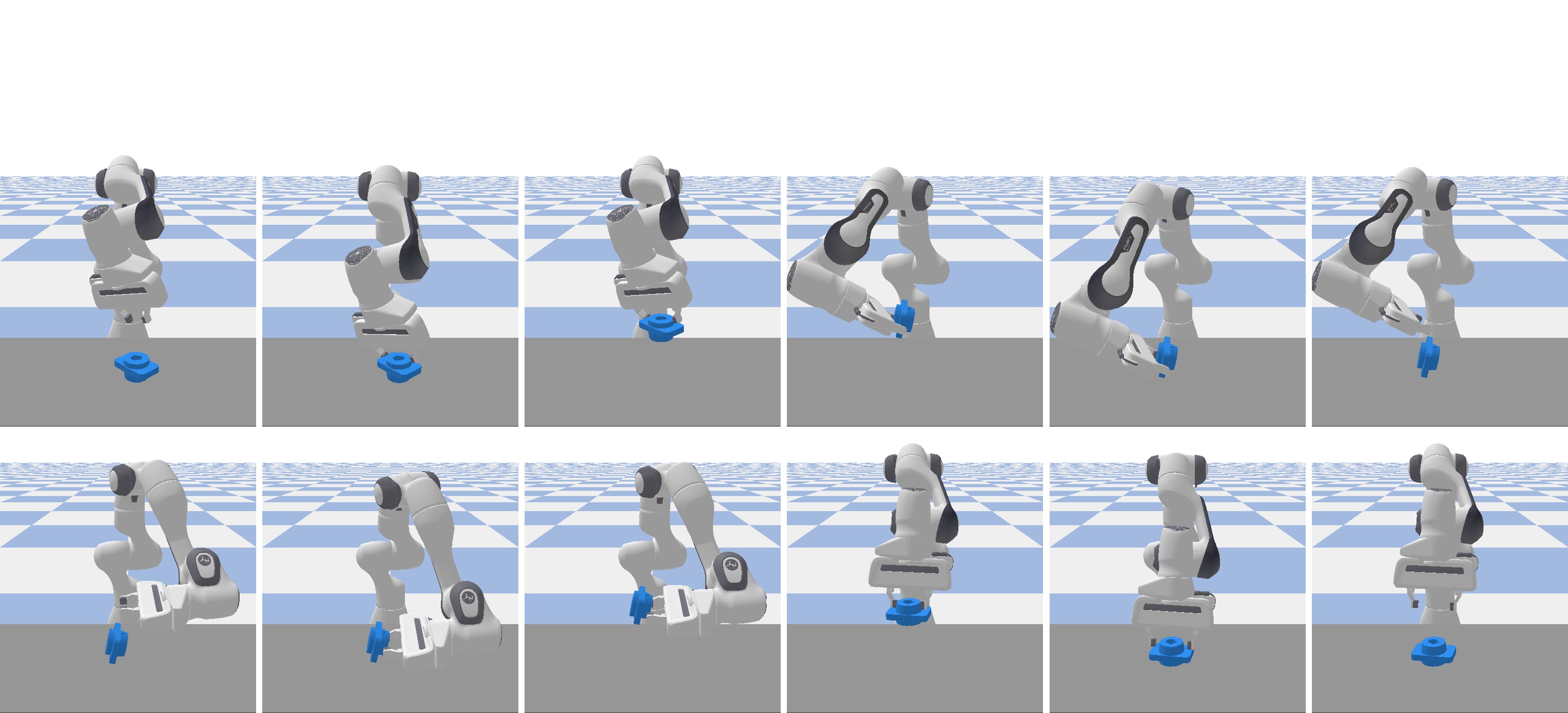}

\includegraphics[width=0.99\linewidth,trim=0 30 0 150,clip,frame]{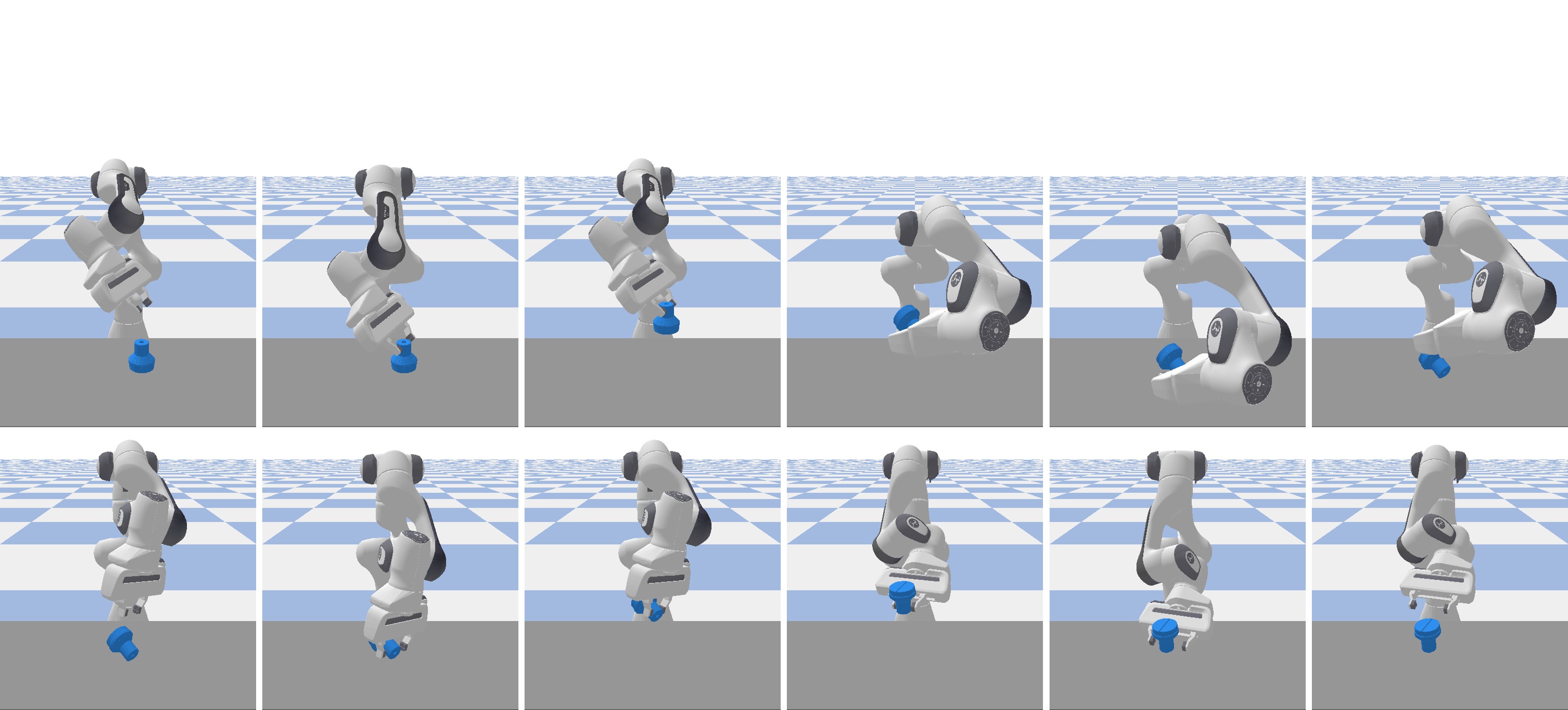}

\includegraphics[width=0.99\linewidth,trim=0 30 0 150,clip,frame]{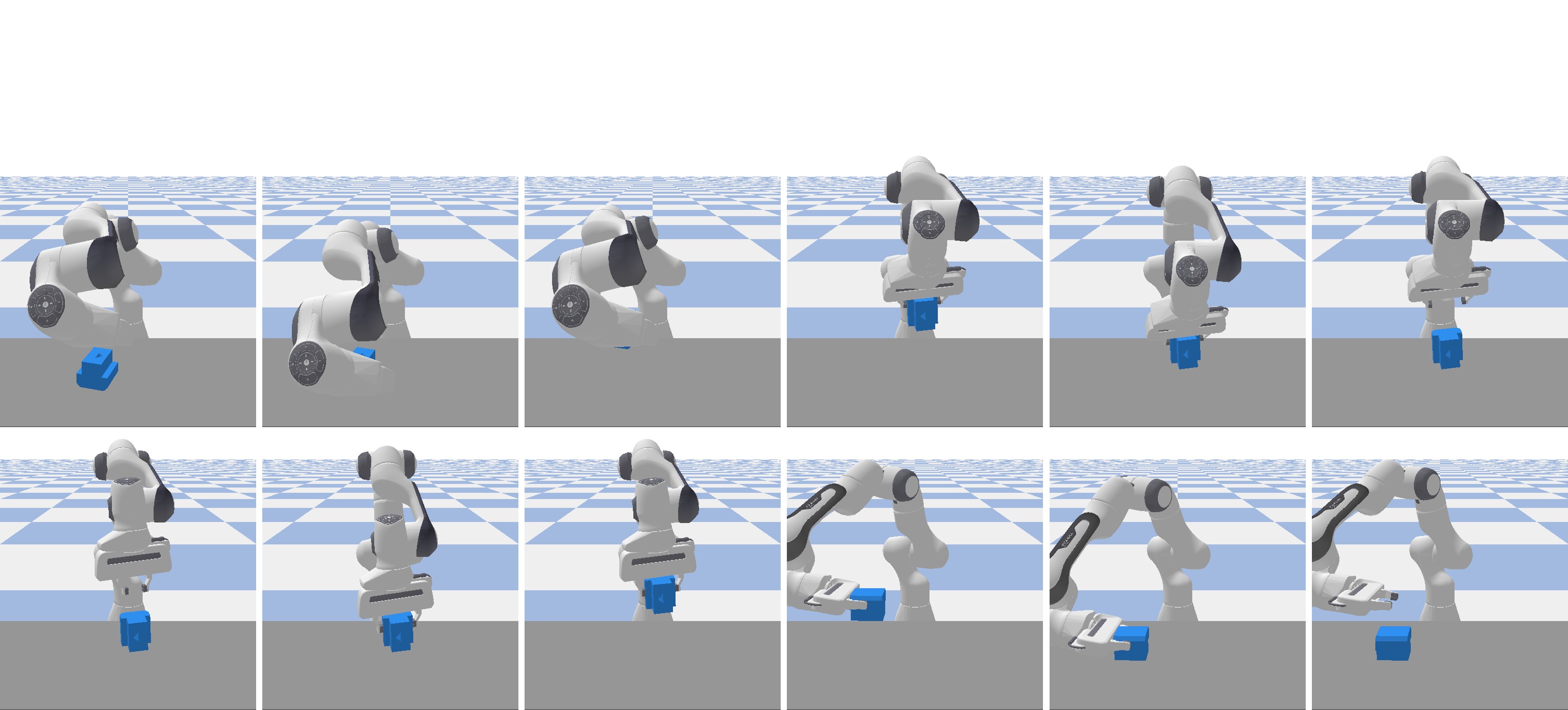}

\caption{Flipping objects with sequential pick-and-place steps. The operation process for each object is in the black border.}

\label{makeup-exps}
\end{figure*}

\begin{figure*}[htbp]
\centering

\includegraphics[width=\linewidth]{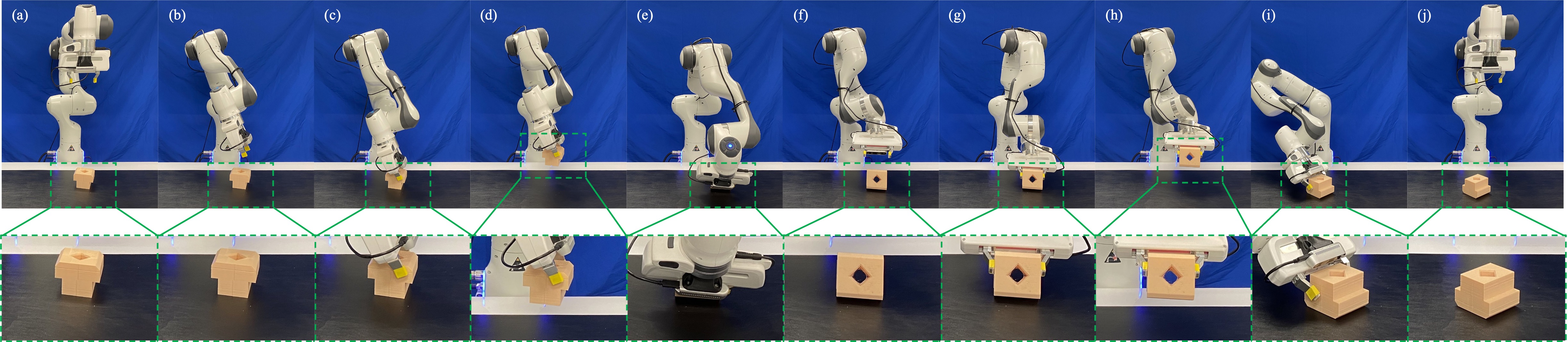}

\caption{An real-robot experiment. (a-e) and (f-j) are two pick-and-place steps. The robot grasps the T-nut in the initial pose (a) and places it in a predicted placement (f). The robot regrasps the T-nut in the placement (f) and places it in the goal placement (j).}

\label{pic_realexp}
\end{figure*}

To demonstrate the performance of our discrimination stage, we compare our approach with an ablated version of our approach and a variation using a discriminator based on PointNet++\cite{qi2017pointnet++} networks. The following approaches are considered for comparison:   

\begin{itemize}
\item $G_{ours} +R_{ours} +D_{PN2}$: In this approach, the discriminator in our approach is replaced by $D_{PN2}$ \cite{paxton2022predicting}, which is a PointNet++ network with abstraction layers. The discriminator $D_{PN2}$ is trained on our training set, using the same data that was used to train our discriminator $D_{ours}$. The score threshold of 0.92, used for selecting stable placements, is consistent with our discriminator $D_{ours}$.

\item $G_{ours} +R _{ours} +D_{rand}$: In this ablated approach, the discriminator in our approach is disabled. The predicted placements after the generation and refinement stages are randomly selected as the final predicted placements for each object.

\end{itemize}

First, we evaluate the generalization performance of the two learning-based discriminators, namely $D_{PN2}$ and $D_{ours}$, on the test set. Fig. \ref{pic_dis_pr} illustrates the precision-recall curves obtained from testing 2,000 collected placements of 99 test objects. These curves show the relationship between the precision of classifying all placements and the recall of stable placements. Notably, our discriminator (the blue curve) demonstrates superior precision and recall compared to $D_{PN2}$ (the green curve).
Next, we compare the results of approaches employing different discrimination stages on the test objects. The accuracy and diversity of predicted placements are reported in the third part of Tab. \ref{table_acc} and \ref{table_div} respectively. Our discriminator exhibits superior accuracy compared to others for nearly all test objects. We hypothesize that incorporating feature manipulation techniques in \cite{jiang2021synergies} facilitates feature extraction for placement classification.
Moreover, our discriminator performs on par with the best results from other approaches in terms of predicting diverse placements for almost all objects. In addition, our approach achieve the highest average diversity of 81.3\% among the aforementioned approaches.

\subsubsection{Comparison with the state-of-the-art}

We compare our approach with a baseline method proposed in \cite{cheng2021learning} that employs a two-stage approach to predict object placements on supporting items. In our evaluation, we utilize plane point clouds as input support points and employ the model parameters provided by the baseline method, which claims to have been trained on a substantial dataset. The predicted placements with the highest scores and the same amount as ours from the baseline method are selected. We conduct a comparative analysis of the accuracy and diversity of the predicted placements between our approach and the baseline. The results of this comparison are presented in the last part of Tab. \ref{table_acc} and \ref{table_div}.

Our approach outperforms the baseline in terms of both average stability and diversity of the predicted placements. Specifically, our approach achieve an average accuracy rate of 90.4\% and an average diversity rate of 81.3\%, compared to the rates of 63.7\% and 33.9\% achieved by the baseline method, respectively. Moreover, Fig. \ref{pic_exp_pred} visualizes the predicted placements of both the baseline method and our approach, which have been verified as stable placements. In the displayed test objects, the stable placements predicted by the baseline method exhibit a lack of diversity, consistent with the results of $G_{L2G}+R_{ours}+D_{ours}$. In contrast, the stable placements predicted by our approach demonstrate both accuracy and diversity.

\subsection{Object Flipping Experiments}

The various predicted placements are ultimately used in extrinsic manipulation tasks. We take object flipping as an example task to illustrate and verify the predicted placements. 
Fig. \ref{makeup-exp0} shows constrained scenarios that require diverse predicted placements for object flipping. In these scenarios, the random initial positions of the objects are given. Goal placements are predicted by our approach. 
However, due to constraints such as object sizes and the fixed base of the robot, shared grasp configurations between the initial and goal placements are unavailable. 
For example, the robot can grasp the bearing and T-nut (the first and last objects in Fig. \ref{makeup-exp0}) in their initial placements. 
However, flipping these objects with a single pick-and-place step is not possible due to their geometric constraints. 
Additionally, due to the fixed robot base, the robot is unable to grasp the knob (the second object in Fig. \ref{makeup-exp0}) in its initial placement. 

The diversity of predicted placements is essential for completing object flipping. Utilizing the predicted placements obtained from our approach, the robot can perform sequential pick-and-place steps to successfully flip novel objects with varying shapes. 
We employ the approach described in Sec. \ref{sec3.4} to compute the grasp configurations between placements with the highest assigned scores across different categories in the predicted results. 
The regrasping operations for different objects are depicted in Fig. \ref{makeup-exps}.
If the diversity of the predicted placements is not of an adequate level, it is difficult to complete the object flipping. 

Finally, we validate our approach on a real platform consisting of a Franka robot arm and a table. The object used in this validation is a 3D-printed object from the test set. 
The extrinsic manipulation scenario, presented in Fig. \ref{pic_realexp}, involves placing the T-nut on the table in a random initial pose. The robot is unable to flip the T-nut with one pick-and-place step.
We use the point cloud fused from multiple views \cite{breyer2021volumetric} as input to our framework to predict placements. Based on the calculated grasp configurations, the robot is able to perform sequential pick-and-place steps to flip the T-nut. 



\section{Conclusion}
In this work, we propose a three-stage framework to generate accurate and diverse stable placements of novel objects based on point clouds. Additionally, we construct a large-scale dataset comprising collected placements and contact information between objects to train proposed neural network models. Comparison experiments demonstrate that our approach outperforms the start-of-the-art, achieving an average accuracy rate of 90.4\% and an average diversity rate of 81.3\% in the generated placements. Moreover, real-robot experiments validate the capability of our approach to accomplish extrinsic manipulation on a supporting plane. 
Feasible and collision-free grasp configurations are crucial, enabling robots to realize the transformation between placements.
To enhance the robot's grasp configurations and motion, we plan to employ learning-based algorithms in future work, aiming to achieve smoother and safer trajectories compared to the current sampling-based algorithm.

\section*{Acknowledgement}

\thanks{The work is supported by the Shenzhen Key Laboratory of Robotics Perception and Intelligence, Southern University of Science and Technology, Shenzhen 518055, China, under Grant ZDSYS20200810171800001.}

\bibliographystyle{ieeetr}
\bibliography{reference}



\end{document}